\documentclass[lettersize,journal]{IEEEtran}
\usepackage{graphicx}
\usepackage{cite}
\usepackage{picinpar}
\usepackage{amsmath}
\usepackage{url}
\usepackage{flushend}
\usepackage[latin1]{inputenc}
\usepackage{colortbl}
\usepackage{soul}
\usepackage{multirow}
\usepackage{pifont}
\usepackage{color}
\usepackage{alltt}
\usepackage[hidelinks]{hyperref}
\usepackage{enumerate}
\usepackage{siunitx}
\usepackage{breakurl}
\usepackage{epstopdf}
\usepackage{pbox}
\usepackage{float}
\usepackage{amssymb}
\usepackage{xcolor}
\usepackage{booktabs}

\usepackage{algorithm}
\usepackage{algorithmic}

\definecolor{Min1}{rgb}{1, 0.75, 0.7}  % 第一名浅红
\definecolor{Min2}{rgb}{1, 0.83, 0.7}   % 第二名浅黄
\definecolor{Min3}{rgb}{1, 0.96, 0.7}  % 第三名浅绿

\begin{document}

\title{\LARGE \bf
RaCalNet: Radar Calibration Network for Sparse-Supervised Dense Depth Estimation
}

\author{Wentao Zhao$^{1}$, Xingrui Qin$^{1}$, Chuan Cao$^{1}$, Yihe Niu$^{2}$, Houcheng Jiang$^{3}$, \\ Tianchen Deng$^{1}$, Rui Guo$^{4}$, Jingchuan Wang$^{1}$$^{*},~\textit{Senior Member, IEEE}$% <-this % stops a space
% \thanks{This work was not supported by any organization}% <-this % stops a space
\thanks{$^{1}$ School of Automation and Intelligent Sensing, Institute of Medical Robotics, Shanghai Jiao Tong University, Shanghai 200240, China. Key Laboratory of System Control and Information Processing, Ministry of Education of China, Shanghai 200240, China.}%
\thanks{$^{2}$ School of Mathematical Sciences, Shanghai Jiao Tong University, Shanghai, China.}%
\thanks{$^{3}$ South China University of Technology, Guangzhou, China.}%
\thanks{$^{4}$ State Grid Intelligence Technology CO., LTD.}%
\thanks{$^{*}$ Corresponding Author: jchwang@sjtu.edu.cn.}%
\thanks{This work is supported by the Technology Project Managed by the State Grid Corporation of China: 5700-202416334A-2-1-ZX.}%

}

% \markboth{IEEE Transactions on Instrumentation and Measurement}%


\maketitle

\begin{abstract}

    Dense depth estimation using millimeter-wave radar typically requires dense LiDAR supervision, generated via multi-frame projection and interpolation, for guiding the learning of accurate depth from sparse radar measurements and RGB images. However, this paradigm is both costly and data-intensive. To address this, we propose RaCalNet, a novel framework that eliminates the need for dense supervision by using sparse LiDAR to supervise the learning of refined radar measurements, resulting in a supervision density of merely around 1\% compared to dense-supervised methods. RaCalNet is composed of two key modules. The Radar Recalibration module performs radar point screening and pixel-wise displacement refinement, producing accurate and reliable depth priors from sparse radar inputs. These priors are then used by the Metric Depth Optimization module, which learns to infer scene-level scale priors and fuses them with monocular depth predictions to achieve metrically accurate outputs. This modular design enhances structural consistency and preserves fine-grained geometric details. Despite relying solely on sparse supervision, RaCalNet produces depth maps with clear object contours and fine-grained textures, demonstrating superior visual quality compared to state-of-the-art dense-supervised methods. Quantitatively, it achieves performance comparable to existing methods on the ZJU-4DRadarCam dataset and yields a 34.89\% RMSE reduction in real-world deployment scenarios. We plan to gradually release the code and models in the future at \url{https://github.com/818slam/RaCalNet.git}.

\end{abstract}


\section{Introduction}

\IEEEPARstart{M}{illimeter-wave} (mmWave) radar has been increasingly adopted in modern measurement systems for its ability to provide direct metric observations under diverse environmental conditions. In particular, radar has enabled effective solutions in diverse tasks such as displacement estimation \cite{liu2023super,ma2023continuous}, simultaneous localization and mapping (SLAM) \cite{li20234d,yang2025compensation}, object detection \cite{jiang2022t,paek2022k}, and gesture recognition \cite{ali2022end,wu2025sparsity}. These capabilities highlight its growing relevance in instrumentation scenarios where accurate and robust perception of physical phenomena is required.

\begin{figure}[!htbp]
    \centering
    \includegraphics[width=0.50\textwidth]{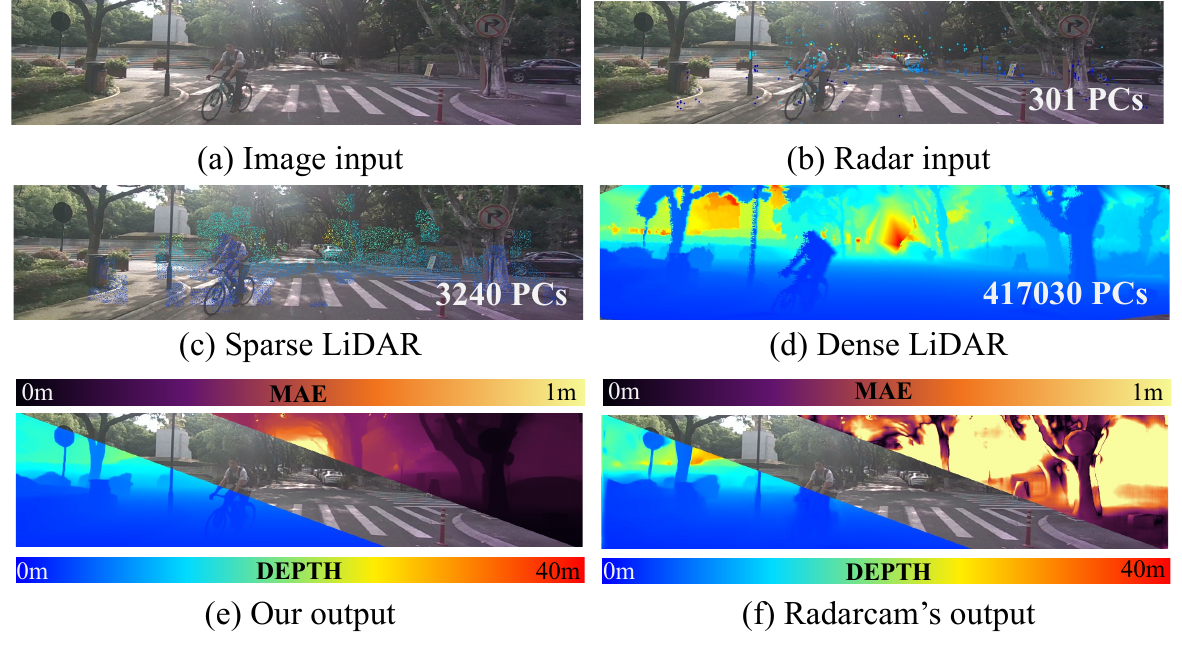}
    \caption{Comparison of depth estimation: 
(a) Image input; 
(b) Radar input (projected on image); 
(c) sparse LiDAR supervision (3,240 points); 
(d) dense LiDAR supervision (417,030 points); 
(e) predicted depth and error map of our method using sparse laser supervision; and 
(f) predicted depth and error map of Radarcam \cite{Radarcam2024} using dense laser supervision.
}
    \label{fig:sample}
\end{figure}

Unlike conventional 3D mmWave radar, which only provides azimuth and range measurements, 4D mmWave radar introduces elevation sensing, resulting in denser point clouds with full spatial information. This richer representation has positioned 4D radar as a promising alternative to LiDAR for various perception tasks. However, mmWave radar point clouds-despite the added vertical dimension-remain extremely sparse and noisy, often two to three orders of magnitude sparser than LiDAR \cite{Long2021}, posing critical challenges for downstream tasks such as dense depth estimation, which is fundamental for 3D reconstruction and autonomous navigation.

To overcome radar sparsity, existing methods often associate radar points with image pixels to create sparse depth priors, which are then densified using learning-based completion networks \cite{Radarcam2024, Singh2023}. To supervise training, they typically synthesize dense labels by aggregating multi-frame LiDAR scans and interpolating missing regions. However, such practices not only incur heavy data collection and processing overhead but also introduce spatial artifacts that degrade depth accuracy and geometric fidelity.

To address these limitations, we propose \textbf{RaCalNet}, a novel radar-guided depth estimation framework that eliminates the need for dense LiDAR supervision by refining sparse radar points into reliable geometric anchors. Specifically, the Radar Recalibration module leverages a cross-modal attention mechanism, trained with single-frame sparse LiDAR, to learn feature correlations between radar and RGB inputs. This module predicts confidence weights for radar points, effectively suppressing noise and refining their pixel-level projections through a lightweight displacement prediction head. 
The resulting refined radar points serve as geometric anchors for the Metric Depth Optimization module, which calibrates monocular depth predictions through a multi-stage process involving global alignment, cluster-wise refinement, and edge-aware smoothing. This module minimizes a geometric alignment loss to enforce consistency between predicted depths and radar geometry, enabling metrically accurate depth estimation aligned with physical measurements.
As shown in Fig.~\ref{fig:sample}, while Radarcam \cite{Radarcam2024} relies on dense LiDAR supervision, our method achieves superior depth estimation using only sparse LiDAR supervision ($\sim$1\% point density), yielding sharper structural boundaries and significantly reduced scale errors.

\begin{figure*}[!htbp] % 注意这里的星号*
    \centering
    \includegraphics[width=0.95\linewidth]{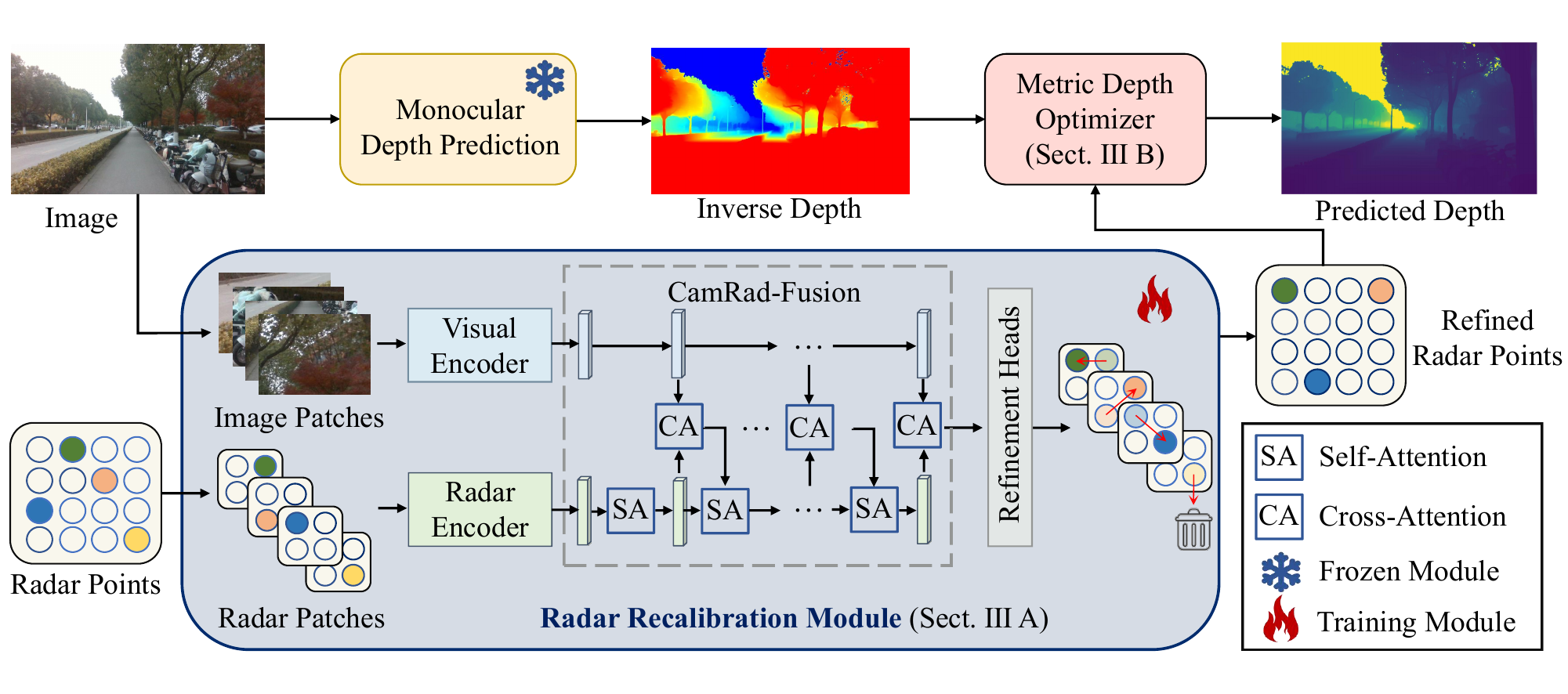} % 适当留白
    \caption{Overview of our proposed RaCalNet framework. The system consists of two main components: (i) a Radar Recalibration module, which performs radar point screening and pixel-wise displacement refinement; and (ii) a Metric Depth Optimizer, which fuses refined radar with monocular depth for metric-accurate output.}
    \label{fig:system}
\end{figure*}

Our key contributions are summarized as follows: (i) We present a novel radar-guided depth estimation framework, RaCalNet, which eliminates the need for dense LiDAR supervision, replacing it with a recalibration and optimization strategy guided by refined radar points; (ii) We leverage a new Radar Recalibration module with cross-modal attention to identify reliable radar points and recalibrate their pixel-level projections, generating sparse yet precise reference anchors for downstream depth prediction; (iii) We design a Metric Depth Optimization module that aligns monocular predictions with refined radar anchors, producing dense depth maps with accurate scale and sharp structural details; (iv) We validate our approach on the public ZJU-4DRadarCam dataset and real-world deployment scenarios, demonstrating significant performance gains over existing methods, including up to 35.30\% RMSE reduction.

\section{Related Work}
\subsection{Depth Estimation with Monocular Vision}

Monocular depth estimation \cite{MiDaS,LeRes,DPT, Depth_anything} inherently suffers from scale ambiguity due to the projective nature of single-view imaging. Recent approaches employ large-scale networks~\cite{MiDaS,LeRes,DPT,Depth_anything} to produce affine-invariant depth maps that capture relative structure but lack absolute metric scale. These models benefit from extensive and diverse datasets to generalize across environments, yet typically require finetuning with external metric supervision~\cite{Wofk2023,CodeVIO,Xie2020} for metric alignment.

To address this, various strategies introduce auxiliary cues. VISLAM~\cite{vislam} performs geometric rectification of camera perspectives, while CodeVIO~\cite{CodeVIO} jointly optimizes visual-inertial odometry (VIO) and dense depth through coupled estimation. Yin et al.~\cite{liu2023unsupervised} proposed an unsupervised monocular depth estimation method that fuses inertial and visual features to produce scale-consistent depth for SLAM systems. More recently, Wofk et al.~\cite{Wofk2023} propose a metric depth estimation framework that globally aligns monocular predictions with sparse but accurate VIO measurements, followed by local optimization. While effective, this approach still relies on tightly coupled VIO systems and may suffer from drift or degraded performance in textureless environments. Yuan et al.~\cite{yuan2025self} proposed a self-supervised monocular depth framework with a depth-motion prior to guide pseudo-LiDAR generation without external calibration. In parallel, a series of video-based methods \cite{Kopf2021,Lee2022,UNO,deng2025neslam,Zhao2023} enhance temporal consistency through learned regularization across frames.

While these methods significantly improve depth prediction quality, they either rely on tightly coupled VIO systems or require temporal sequences and large-scale supervision, limiting their applicability in unconstrained or dynamic environments. In contrast, our goal is to improve monocular depth accuracy by introducing sparse yet metrically accurate millimeter-wave radar as geometric anchors.

\subsection{Depth Estimation from Radar-Camera Fusion}

Fusing radar and camera data for metric depth estimation has emerged as a promising direction due to radar's robustness to lighting and weather conditions. However, the inherent sparsity and noise of radar measurements present significant challenges for generating dense and accurate depth maps.

To address this, prior works have explored various fusion strategies. Lin et al.~\cite{Lin2020} introduced a two-stage encoder--decoder pipeline to denoise radar point clouds before fusing them with image features. Lo et al.~\cite{Lo2021} proposed height-extended radar representations to enhance spatial context. Long et al.~\cite{Long2021} designed a Radar-to-Pixel (R2P) network that leverages Doppler velocity and optical flow to establish pixel-level correspondences between radar and image domains. Their follow-up work~\cite{Long2022} integrated multi-sweep radar accumulation with dense LiDAR supervision in an image-guided completion framework. Cui et al.~\cite{cui2022dense} fused event camera data with sparse depth using density-based point clustering and surface reconstruction to generate dense depth maps.

Several recent methods aim to enhance fusion through confidence modeling or temporal aggregation. R4dyn~\cite{R4dyn2021} treats radar points as weak supervision but focuses narrowly on dynamic vehicles. Singh et al.~\cite{Singh2023} proposed a gated fusion approach with confidence-aware radar--image correspondences, yet their method relies on accumulating up to 161 LiDAR frames for training. Radarcam~\cite{Radarcam2024} combines raw radar data with monocular predictions to produce dense depth maps, but also fundamentally depends on dense LiDAR supervision during training.

Despite these advances, existing methods still suffer from two key limitations: (1) they heavily rely on dense LiDAR supervision, often requiring multi-frame accumulation and interpolation; (2) they perform direct matching or regression from noisy radar points to image regions, leading to unreliable priors. In contrast, we propose to recalibrate sparse radar signals through guided refinement, enabling accurate metric-scale estimation with minimal supervision.

\section{Method}
In this work, we address the task of dense, metrically accurate depth estimation \(\hat{d} \in \mathbb{R}_+^{H \times W}\) by fusing monocular RGB images \(I \in \mathbb{R}^{3 \times H \times W}\) with sparse radar data \(P = \{p_i \in \mathbb{R}^3\}_{i=1}^k\). 
The overall architecture of our RaCalNet is illustrated in Fig.~\ref{fig:system}. Our pipeline first recalibrates and refines radar measurements using sparse LiDAR supervision in Radar Recalibration module, producing accurate depth anchors. These refined anchors are then used to optimize monocular depth predictions, specifically to correct their scale and enhance their structural consistency, resulting in accurate metric-scale estimation. Each component of our framework is detailed in the following subsections.

\subsection{Radar Recalibration Module}

Due to multipath reflections, sensor noise, and imperfect extrinsic calibration, the direct projection of radar points onto the image plane can result in substantial depth errors. Therefore, in addition to filtering reliable radar measurements, it is crucial to recalibrate and refine projected points within localized image regions to account for systemic errors and outliers. To achieve this, we adopt a patch-based processing strategy: for each radar point projected onto the image plane, we extract a corresponding image patch centered at its location, forming a pair of radar and image patches. These point-wise pairs are independently encoded and fused to enable fine-grained refinement.

1) Network Architecture: Unlike conventional approaches that construct a semi-dense radar depth map, our Radar Recalibration module predicts both a confidence score and a pixel-wise displacement for each projected radar point. This allows the network to directly assess reliability and spatial alignment.

Motivated by recent advances in sparse 3D representation and cross-modal point processing~\cite{wang2024sfpnet,wang2025salt}, our Radar Recalibration module employs a cross-modal fusion design. The image encoder is based on ResNet-34~\cite{resnet34}, producing hierarchical feature maps with channel dimensions of 32, 64, 128, 256, and 512. These multi-scale features are aggregated using a Feature Pyramid Network (FPN), ensuring the preservation of spatial detail and semantic richness. To further enhance feature alignment and interaction, we incorporate channel and spatial attention mechanisms, which help highlight salient structures and suppress irrelevant background noise in a content-adaptive manner.

The radar encoder is implemented as a multi-layer perceptron (MLP) with fully connected layers of dimensions 32, 64, 128, 256, and 512. To facilitate effective radar-image fusion, image features $\mathbf{F}_{img} \in \mathbb{R}^{B \times K \times D \times H \times W}$ are reshaped along the spatial dimensions to $\mathbb{R}^{B \times K \times D \times N_{img}}$ with $N_{img} = H \times W$, so as to align with radar features $\mathbf{F}_{rad} \in \mathbb{R}^{B \times K \times D}$, where $B$ is the batch size, $K$ is the number of radar points, and $D$ is the feature dimension. Both modalities are then flattened along the spatial axis and passed through stacked attention layers, forming the CamRad-Fusion net for effective cross-modal interaction and feature enhancement.

For each sample, radar features are represented as $\mathbf{X}_{rad} \in \mathbb{R}^{K \times D}$. The self-attention (SA) operation is defined as:
\begin{equation}
\mathrm{SA}(\mathbf{X}_{rad}) = 
\mathrm{softmax} \left( \frac{\mathbf{Q} \mathbf{K}^\top}{\sqrt{d_k}} \right) \mathbf{V} \mathbf{W}_O,
\tag{1}
\end{equation}
where the query, key, and value matrices are computed as $\mathbf{Q} = \mathbf{X}_{rad} \mathbf{W}_Q$, $\mathbf{K} = \mathbf{X}_{rad} \mathbf{W}_K$, and $\mathbf{V} = \mathbf{X}_{rad} \mathbf{W}_V$, with learnable projections $\mathbf{W}_Q, \mathbf{W}_K, \mathbf{W}_V \in \mathbb{R}^{D \times d_k}$ and $\mathbf{W}_O \in \mathbb{R}^{d_k \times D}$. This enables each radar point to attend to others and capture global contextual dependencies.

To incorporate visual context, cross-attention (CA) is applied with radar features as queries and image features as keys and values. For each radar point, the corresponding visual patch features are reshaped to $\mathbf{X}_{img} \in \mathbb{R}^{D \times N_{img}}$ and radar features are viewed as $\mathbf{X}_{rad} \in \mathbb{R}^{D \times 1}$, and the attention is computed as:
\begin{equation}
\mathrm{CA}(\mathbf{X}_{rad}, \mathbf{X}_{img}) = 
\mathrm{softmax} \left( \frac{\mathbf{Q} \mathbf{K}^\top}{\sqrt{d_k}} \right) \mathbf{V} \mathbf{W}_O,
\tag{2}
\end{equation}
where $\mathbf{Q} = \mathbf{X}_{rad} \mathbf{W}_Q$, $\mathbf{K} = \mathbf{X}_{img} \mathbf{W}_K$, and $\mathbf{V} = \mathbf{X}_{img} \mathbf{W}_V$. The output preserves the radar feature resolution and is reshaped back to $\mathbb{R}^{B \times K \times D}$ for subsequent processing. This cross-modal attention allows the network to enhance radar representations with informative visual cues. Finally, the refined radar features are fed into the refinement heads, which consist of a two-layer MLP that separately predicts a confidence score and a pixel-wise displacement for each radar point.

2) Confidence and Displacement Prediction: Given a radar point \( p_i \) and its corresponding image patch \( \mathcal{C}_i \in \mathbb{R}^{3 \times H \times W} \) around the projection location, RaCalNet \( h_\theta \) is designed with two output branches: a confidence prediction head \( h_\theta^{\mathrm{conf}} \) that estimates point-wise reliability, and a displacement prediction head \( h_\theta^{\mathrm{disp}} \) that predicts pixel-level displacement. The network outputs are defined as:
\[
\hat{\mathbf{y}}_i = h_\theta^{\mathrm{conf}}(\mathcal{C}_i, p_i) \in [0,1], \quad 
\hat{\mathbf{o}}_i = h_\theta^{\mathrm{disp}}(\mathcal{C}_i, p_i) \in \mathbb{R}^2,
\]
where \( \hat{\mathbf{y}}_i \) represents the predicted confidence score indicating the correspondence reliability between radar point \( p_i \) and its projected pixel, while \( \hat{\mathbf{o}}_i = (\Delta u, \Delta v) \) is the estimated 2D displacement vector for refining the projected location.

With \(k\) radar points in the cloud \(P\), these predictions guide the reliability screening and spatial refinement process:
\begin{equation}
(u_i^{\text{ref}}, v_i^{\text{ref}}) =
\begin{cases}
(u_i, v_i) + \hat{\mathbf{o}}_i, & \text{if } \hat{\mathbf{y}}_i \geq \tau \\
\text{discarded}, & \text{otherwise}
\end{cases}
\tag{3}
\end{equation}
Here, $(u_i, v_i)$ is the original 2D projection of radar point $p_i$, and 
$\hat{\mathbf{o}}_i = (\Delta u, \Delta v)$ is the predicted displacement. 
The refined coordinate $(u_i^{\text{ref}}, v_i^{\text{ref}})$ is used only if 
the predicted confidence $\hat{\mathbf{y}}_i$ exceeds a predefined threshold $\tau$; 
otherwise, the point is discarded.

3) Training Loss: The network is trained using sparse LiDAR points projected onto the image plane as depth ground truth, denoted as $d_{gt}(x, y)$ for valid LiDAR pixel locations. Dual learning objectives are defined over the radar points $p \in P$ for confidence classification and displacement regression.

\textit{Confidence Labels and Loss:} For each radar point $p$ with projected pixel location $(u_p, v_p)$, we define a local neighborhood $\mathcal{R}(p)$ (e.g., $5 \times 5$ pixels). The binary confidence label $y^{(conf)}_p \in \{0,1\}$ is assigned based on the count of valid LiDAR points within $\mathcal{R}(p)$ that meet the depth consistency criterion:
\begin{equation}
y^{(conf)}_p = \begin{cases}
1 & \text{if } |\{ x \in \mathcal{R}(p) \mid d_{gt}(x) > 0 \\
& \quad \text{and } |d_{gt}(x) - d(p)| < \tau(p) \}| \geq \eta \\
0 & \text{otherwise}
\end{cases}
\tag{4}
\end{equation}
where $\eta$ is a minimum count threshold (e.g., $\eta = 3$), and the adaptive depth error threshold $\tau(p)$ is defined based on the radar point's measured range $d(p)$:
\begin{equation}
\tau(p) = \begin{cases}
0.5\,\mathrm{m} & d(p) < 30\,\mathrm{m} \\
0.75\,\mathrm{m} & 30\,\mathrm{m} \leq d(p) \leq 50\,\mathrm{m} \\
1.0\,\mathrm{m} & d(p) > 50\,\mathrm{m}
\end{cases}
\tag{5}
\end{equation}
The confidence classification loss is the binary cross-entropy over all radar points with valid LiDAR ground truth available in their neighborhood:
\begin{equation}
\begin{split}
\mathcal{L}_{conf} = \frac{1}{|P_{valid}|} \sum_{p \in P_{valid}} 
&-\biggl[ y^{(conf)}_p \log \hat{y}^{(conf)}_p \\
&+ (1 - y^{(conf)}_p) \log(1 - \hat{y}^{(conf)}_p) \biggr]
\end{split}
\tag{6}
\end{equation}
Here, \( \hat{y}^{(conf)}_p = h_\theta^{\mathrm{conf}}(\mathcal{C}_i, p) \) is the predicted confidence, and $P_{valid} = \{ p \in P \mid \exists x \in \mathcal{R}(p), d_{gt}(x) > 0 \}$ is the set of radar points where LiDAR ground truth exists within their defined local neighborhood for label assignment.

\textit{Displacement Labels and Loss:} For each radar point $p$, we search within a larger local $h \times w$ neighborhood $\mathcal{N}(p)$ centered on its projection $(u_p, v_p)$. A smaller sliding window of size $h^* \times w^*$ is convolved over $\mathcal{N}(p)$. The displacement label $y^{(disp)}_p$ is determined by the offset from $(u_p, v_p)$ to the center of the window position $\mathbf{c}^*$ that contains the maximum number of valid LiDAR points satisfying the depth consistency criterion:
\begin{equation}
\begin{split}
\mathbf{c}^* = \underset{\mathbf{c} \in \mathcal{N}(p)}{\arg\max} 
\Bigl| \bigl\{ x \in \mathcal{W}_{\mathbf{c}} 
&\mid d_{gt}(x) > 0 \\
&\quad \text{and } |d_{gt}(x) - d(p)| < \tau(p) \bigr\} \Bigr|
\end{split}
\tag{7}
\end{equation}
\begin{equation}
y^{(disp)}_p = \mathbf{c}^* - (u_p, v_p)
\tag{8}
\end{equation}
where $\mathcal{W}_{\mathbf{c}}$ denotes the set of pixels within the sliding window centered at candidate location $\mathbf{c}$. The predicted pixel displacement $\hat{y}^{(disp)}_p$ is supervised using the Smooth L1 loss:
\begin{equation}
\mathcal{L}_{disp} = \frac{1}{|P_{valid}|} \sum_{p \in P_{valid}} 
\mathrm{SmoothL1}\left( \hat{y}^{(disp)}_p, y^{(disp)}_p \right)
\tag{9}
\end{equation}
Here, \( \hat{y}^{(disp)}_p = h_\theta^{\mathrm{disp}}(\mathcal{C}_i, p) \). The final training loss combines both objectives:
\begin{equation}
\mathcal{L}_{total} = \lambda_{conf} \mathcal{L}_{conf} + \lambda_{disp} \mathcal{L}_{disp}
\tag{10}
\end{equation}

\subsection{Metric Depth Optimization}

We now recover the final metrically scaled depth map by aligning the scale-ambiguous monocular inverse depth predictions with radar-based metric measurements. A frozen monocular depth estimation network~\cite{Depth_anything_v2} is employed to generate dense, scale-ambiguous inverse depth maps \(\hat{d}_m \in \mathbb{R}_+^{H \times W}\). This transformer-based model~\cite{Transformers}, trained using scale-invariant losses over diverse datasets, preserves fine-grained textures and sharp geometric boundaries, which are essential for downstream metric refinement.

\begin{algorithm}[t]
\caption{Metric Depth Optimization}
\label{alg:refined_metric_depth}
\begin{algorithmic}[1]
\REQUIRE Monocular inverse depth map $\hat{d}_m$, radar point set $\{(u_i, v_i, d_{\text{radar}}^{(i)})\}$, threshold candidates $\mathcal{T}_d$, regularization weight $\lambda$
\ENSURE Refined metric depth $\hat{d}_{\text{final}}$

\vspace{0.2em}
\STATE \textbf{Stage 1: Global Affine Alignment}
\FORALL{$t \in \mathcal{T}_d$}
  \STATE $\mathcal{S}_t \gets \{ i \mid 0 < \hat{d}_m(u_i, v_i) < t \}$ \COMMENT{Filter unreliable predictions}
  \STATE $X \gets [\hat{d}_m(u_i, v_i), \mathbf{1}]_{i \in \mathcal{S}_t}$, $Y \gets [1/d_{\text{radar}}^{(i)}]_{i \in \mathcal{S}_t}$
  \STATE Solve $\min_\theta \|Y - X\theta\|^2 + \lambda\|\theta\|^2$, obtain $\theta_t = (\alpha_t, \beta_t)$
  \STATE Compute relative error: $\mathcal{L}_t = \frac{1}{|\mathcal{S}_t|} \sum_{i \in \mathcal{S}_t} \frac{|d_{\text{pred}}^{(i)} - d_{\text{radar}}^{(i)}|}{d_{\text{radar}}^{(i)}}$
  \IF{$\mathcal{L}_t < \mathcal{L}^*$}
    \STATE $T^* \gets t$, $\theta^* \gets \theta_t$
  \ENDIF
\ENDFOR
\STATE Apply global alignment: $\hat{d}_{\text{global}}(u,v) \gets 1 / (\alpha^* \cdot \hat{d}_m(u,v) + \beta^*)$

\vspace{0.4em}
\STATE \textbf{Stage 2: Cluster-wise Local Refinement}
\STATE Extract clustering features: \\
\hspace{\algorithmicindent} $\mathbf{f}(u,v) = [u/W, v/H, \log(\hat{d}_{\text{global}}(u,v))]$
\STATE Determine number of clusters: \\
\hspace{\algorithmicindent} $k \gets \min\left(8, \max\left(2, \left\lfloor N_{\text{valid}}/500 \right\rfloor\right)\right)$
\STATE Cluster valid pixels via K-means: $\mathcal{C} \gets \{\mathcal{C}_1, ..., \mathcal{C}_k\}$
\FORALL{cluster $\mathcal{C}_c$}
  \STATE Identify radar samples within cluster: $\mathcal{S}_c \gets \{ i \mid (u_i, v_i) \in \mathcal{C}_c \}$
  \IF{$|\mathcal{S}_c| \geq 3$}
    \STATE $X_c \gets [\hat{d}_m(u_i, v_i), \mathbf{1}]_{i \in \mathcal{S}_c}$, $Y_c \gets [1/d_{\text{radar}}^{(i)}]_{i \in \mathcal{S}_c}$
    \STATE Solve $\min_{\theta_c} \|Y_c - X_c\theta_c\|^2$, obtain $(\alpha_c, \beta_c)$
    \STATE Update: $\hat{d}_{\text{local}}^{(c)}(u,v) \gets 1 / (\alpha_c \cdot \hat{d}_m(u,v) + \beta_c), \forall (u,v) \in \mathcal{C}_c$
  \ELSE
    \STATE Use global prediction: $\hat{d}_{\text{local}}^{(c)} \gets \hat{d}_{\text{global}}$
  \ENDIF
\ENDFOR

\vspace{0.4em}
\STATE \textbf{Stage 3: Edge-Aware Smoothing}
\STATE Compute gradient magnitude: $E(u,v) \gets \|\nabla \hat{d}_{\text{global}}(u,v)\|$
\STATE Define edge mask: $\mathcal{M}_{\text{edge}} \gets \{ (u,v) \mid E(u,v) > \tau \}$
\STATE Apply Gaussian smoothing: $\hat{d}_{\text{smooth}} \gets \text{GaussianBlur}(\hat{d}_{\text{local}}, \sigma=1.5, \text{kernel}=5\times5)$
\STATE Fuse results:
\[
\hat{d}_{\text{final}}(u,v) \gets 
\begin{cases} 
\hat{d}_{\text{local}}(u,v), & (u,v) \in \mathcal{M}_{\text{edge}} \\
\hat{d}_{\text{smooth}}(u,v), & \text{otherwise}
\end{cases}
\]

\RETURN $\hat{d}_{\text{final}}$
\end{algorithmic}
\end{algorithm}

To achieve robust alignment with sparse radar measurements, we propose a multi-stage optimization strategy that progressively refines the scale and bias in both global and local contexts. Let $d_{\text{radar}}^{(i)}$ denote the depth of a refined radar point $p_i$, and $(u_i, v_i)$ its image coordinates. Our approach includes three key components:

1) Global Affine Alignment: We begin by estimating a global affine transformation in the inverse depth space. Given predicted inverse depths $\hat{d}_m(u_i, v_i)$ and ground-truth inverse depths $1 / d_{\text{radar}}^{(i)}$, we solve the following regularized least squares problem:
\begin{equation}
\min_{\alpha, \beta} \sum_{i \in \mathcal{S}} \left( \frac{1}{d_{\text{radar}}^{(i)}} - \alpha \cdot \hat{d}_m(u_i, v_i) - \beta \right)^2 + \lambda (\alpha^2 + \beta^2)
\tag{11}
\end{equation}
where $\mathcal{S}$ denotes the set of valid radar-projected pixels. To enhance robustness, we dynamically search for the optimal inverse depth threshold $T^*$ to filter out unreliable predictions:
\begin{equation}
\mathcal{S}_t = \left\{ i \mid 0 < \hat{d}_m(u_i, v_i) < t \right\}, \quad T^* = \arg\min_{t} \mathcal{L}(t)
\tag{12}
\end{equation}
where $\mathcal{L}(t)$ is the relative alignment error under threshold $t$.

2) Depth-Aware Local Refinement: After global alignment, we further refine the depth map by leveraging local geometric consistency. We perform unsupervised K-means clustering on valid pixels using normalized spatial coordinates and logarithmic aligned depth:
\begin{equation}
\mathbf{f}(u, v) = \left[ \frac{u}{H}, \frac{v}{W}, \log(\hat{d}_{\text{global}}(u,v)) \right]
\tag{13}
\end{equation}
Each cluster is assumed to represent a locally planar or smoothly varying region. Within each cluster, we identify radar points and estimate a local affine transformation $(\alpha_c, \beta_c)$ via least squares, then update depth values:
\begin{equation}
\hat{d}_c(u, v) = \frac{1}{\alpha_c \cdot \hat{d}_m(u,v) + \beta_c}, \quad \forall (u,v) \in \text{cluster } c
\tag{14}
\end{equation}

3) Edge-Aware Smoothing: To enforce spatial smoothness while preserving depth discontinuities, we compute edge maps using Sobel gradients of the aligned depth. A Gaussian filter is then selectively applied to non-edge regions, suppressing high-frequency noise in homogeneous areas without blurring object boundaries.

Algorithm~\ref{alg:refined_metric_depth} summarizes the full optimization process. This multi-stage alignment improves both the metric accuracy and geometric coherence of monocular depth predictions.

\section{Experiments}

To evaluate the effectiveness and transferability of RaCalNet, we conduct experiments on both a public dataset and a real-world deployment scenario. Our primary benchmark is the ZJU-4DRadarCam dataset~\cite{Radarcam2024}, which offers time-synchronized 4D radar, RGB, and LiDAR data across diverse environments. In contrast, widely used datasets such as nuScenes~\cite{nuscenes} rely on 3D radar with only azimuth-range measurements on a fixed horizontal plane, lacking elevation information and thus unsuitable for evaluating 4D radar-based methods. To further assess real-world adaptability, we collect a proprietary dataset using a robotic platform equipped with production-grade sensors. Featuring distinct hardware configurations and capturing campus navigation scenarios, this real-world dataset allows for a more comprehensive evaluation of RaCalNet in practical applications.

For comparison, we adopt Radarcam~\cite{Radarcam2024} as our main baseline, as it represents a recent state-of-the-art method in radar-guided dense depth estimation. Radarcam leverages dense LiDAR supervision and multi-frame fusion, providing strong performance on their ZJU-4DRadarCam dataset. Its open-source implementation and methodological relevance to our approach make it a suitable and competitive reference point.

\subsection{Datasets}

1) ZJU-4DRadarCam Dataset~\cite{Radarcam2024}: This public dataset was collected using a ground robotic platform equipped with an Oculii EAGLE 4D radar, a RealSense D455 camera, and a RoboSense M1 LiDAR. It includes 33,409 synchronized radar-camera keyframes, with 29,312 used for training and validation, and 4,097 reserved for testing. The dataset captures diverse scenes spanning both urban roads and off-road trails.

\begin{figure}[!tbp]
    \centering
    \includegraphics[width=0.3\textwidth]{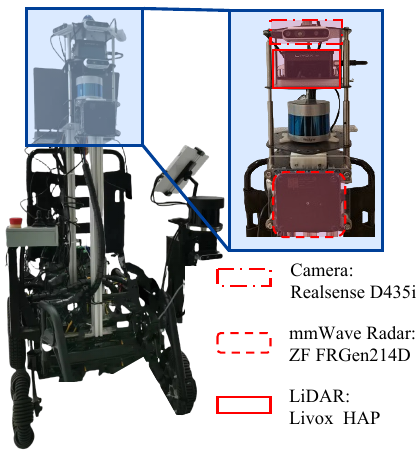}
    \caption{Data collection platform and sensors. 
(Left) Overall view of the ground robotic platform used for real-world data collection. 
(Right) Close-up views of the multimodal sensors: RealSense D435I camera (top), Livox HAP LiDAR (middle), and ZF FRGen21 4D radar (bottom).}
    \label{fig:model}
\end{figure}

2) Real-World Deployment Scenarios: To further validate the generalizability of our approach, we collected a proprietary multimodal dataset using a different robotic platform equipped with a ZF FRGen21 4D radar, a RealSense D435I camera, and a Livox HAP LiDAR, which provides the ground-truth depth labels (see Fig.~\ref{fig:model}). Compared to the ZJU setup, this platform features different radar characteristics (e.g., field of view) and a LiDAR sensor with a unique scanning pattern. The dataset consists of 22,423 synchronized keyframes, primarily collected in campus navigation scenarios that emphasize dynamic pedestrians and varying lighting conditions. Among them, 20,156 frames are used for training and validation, and 2,267 for testing.

\subsection{Implementation Details and Evaluation Protocol}

The encoder backbone was initialized with ImageNet pre-trained weights~\cite{imagenet}, while all other layers were randomly initialized. For the ZJU-4DRadarCam dataset, input images were resized and cropped to $352 \times 640$, with a fixed region-of-interest (ROI) size of $35 \times 35$. We employed the Adam optimizer ($\beta_1$=0.9, $\beta_2$=0.999) with an initial learning rate of $2 \times 10^{-4}$, decayed step-wise over 130 epochs to a final rate of $1 \times 10^{-6}$. The same data augmentation strategies were applied to both datasets: random horizontal flipping (probability = 0.5), and random variations in saturation, brightness, and contrast.

For real-world deployment scenarios, input images were resized to $224 \times 416$ while maintaining the same ROI configuration. The optimizer and hyperparameter settings remained identical, except for a longer training schedule of 150 epochs. All models were trained on an NVIDIA RTX 3090 GPU with a batch size of 28, and the full training process for RaCalNet took approximately 24 hours.

For quantitative evaluation of depth estimation performance, we adopt standard metrics widely used in prior work \cite{Neuralrecon}, including Mean Absolute Error (MAE), Root Mean Square Error (RMSE), Absolute Relative Error (AbsRel), Squared Relative Error (SqRel), and thresholded accuracy ($\delta_1$). All metrics are computed separately within three evaluation ranges: 0-50\,m, 0-70\,m, and 0-80\,m, to provide a comprehensive assessment across varying depths.

\subsection{Evaluation on ZJU-4DRadarCam}

\begin{figure}[!tbp]
    \centering
    \includegraphics[width=\linewidth]{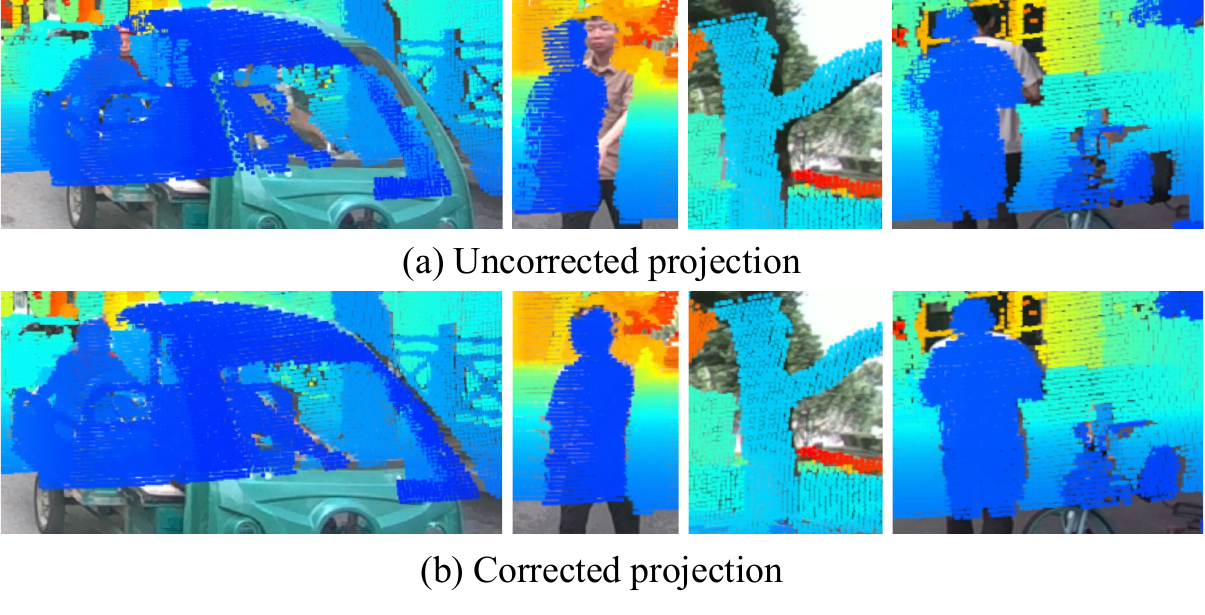}
    % 适当留白
    \caption{Visualization of depth inaccuracies in the ZJU-4DRadarCam dataset. The top row illustrates misalignment in the LiDAR depth map (the original ground truth). The bottom row shows our calibrated version of the ground truth, where the depth projection exhibits significantly improved accuracy.}
    \label{fig:zju_error}
\end{figure}

We initially evaluated our method and all baselines using the original ZJU-4DRadarCam dataset, strictly following the settings reported in Radarcam, where multi-frame densified LiDAR supervision is used. Under this configuration, our RaCalNet employed only sparse single-frame LiDAR supervision, while the comparative methods were trained with their default settings. Quantitative results under this setting are reported in Table~\ref{tab:zju_original}. While RaCalNet performed competitively, it did not exhibit significant improvements over the Radarcam (DPT+Var+RC-Net), which was contrary to our expectations given the design of our framework. Notably, our method produced visibly cleaner and more geometrically consistent depth maps, but these qualitative improvements were not fully captured by the reported metrics. This discrepancy prompted us to re-examine the ground-truth depth used for evaluation. We found that the dataset suffered from imperfect extrinsic calibration, resulting in noticeable misalignment between LiDAR point clouds and RGB images. Fig.~\ref{fig:zju_error} illustrates these geometric inconsistencies, which likely degraded the effectiveness of evaluation.

\begin{table}[!tbp]
\centering
\caption{EVALUATIONS ON ZJU-4DRADARCAM (mm)}
\label{tab:zju_original}
\small
\setlength{\tabcolsep}{5pt}
\resizebox{\columnwidth}{!}{%
\begin{tabular}{@{}llccccccc@{}}
\toprule
Dist & Method & MAE$\downarrow$ & RMSE$\downarrow$ & AbsRel$\downarrow$ & SqRel$\downarrow$ & $\delta_1\uparrow$ \\
\midrule
\multirow{4}{*}{50m} 
& DORN \cite{Lo2021} & 2210.171 & 4129.691 & 0.157 & 939.348 & 0.783 \\
& Singh \cite{Singh2023} & 1785.391 & 3704.636 & 0.146 & 966.133 & 0.831 \\
& Radarcam \cite{Radarcam2024} & \textbf{1067.531} & 2817.362 & \textbf{0.087} & 575.838 & \textbf{0.922} \\
& Ours & 1328.045 & \textbf{2248.606} & 0.095 & \textbf{340.544} & 0.893  \\
\midrule

\multirow{4}{*}{70m}
& DORN \cite{Lo2021} & 2402.180 & 4625.231 & 0.160 & 1021.805 & 0.777 \\
& Singh \cite{Singh2023} & 1932.690 & 4137.143 & 0.147 & 1014.454 & 0.828 \\
& Radarcam \cite{Radarcam2024} & \textbf{1157.014} & 3117.721 & \textbf{0.087} & 601.052 & \textbf{0.921} \\
& Ours & 1492.442 & \textbf{2576.835} & 0.100 & \textbf{410.569} & 0.882 \\
\midrule

\multirow{4}{*}{80m}
& DORN \cite{Lo2021} & 2447.571 & 4760.016 & 0.161 & 1038.919 & 0.776 \\
& Singh \cite{Singh2023} & 1979.459 & 4309.314 & 0.147 & 1034.148 & 0.828 \\
& Radarcam \cite{Radarcam2024} & \textbf{1183.471} & 3228.999 & \textbf{0.088} & 610.501 & \textbf{0.920} \\
& Ours & 1531.343 & \textbf{2659.084} & 0.101 & \textbf{428.687} & 0.880 \\
\bottomrule
\end{tabular}%
}
\end{table}

To address this issue, we refined the sensor alignment by optimizing the extrinsic parameters and reprojecting the LiDAR data, which significantly improved the spatial consistency between LiDAR and camera views. We then re-trained both our method and the strongest-performing baseline on the corrected dataset using the same training configurations as before. Notably, RaCalNet continued to rely only on sparse single-frame LiDAR, while the baseline retained their original dense supervision.

After correction, overall performance increased, and RaCalNet achieved state-of-the-art results. As shown in Table~\ref{tab:zju_corrected}, our method achieved RMSE reductions of 35.08\%, 29.12\%, and 27.27\% at 50m, 70m, and 80m respectively, compared to the strongest baseline. These results validate the effectiveness of both the correction process and our overall approach.

\begin{figure}[!tbp]
    \centering
    \includegraphics[width=\linewidth]{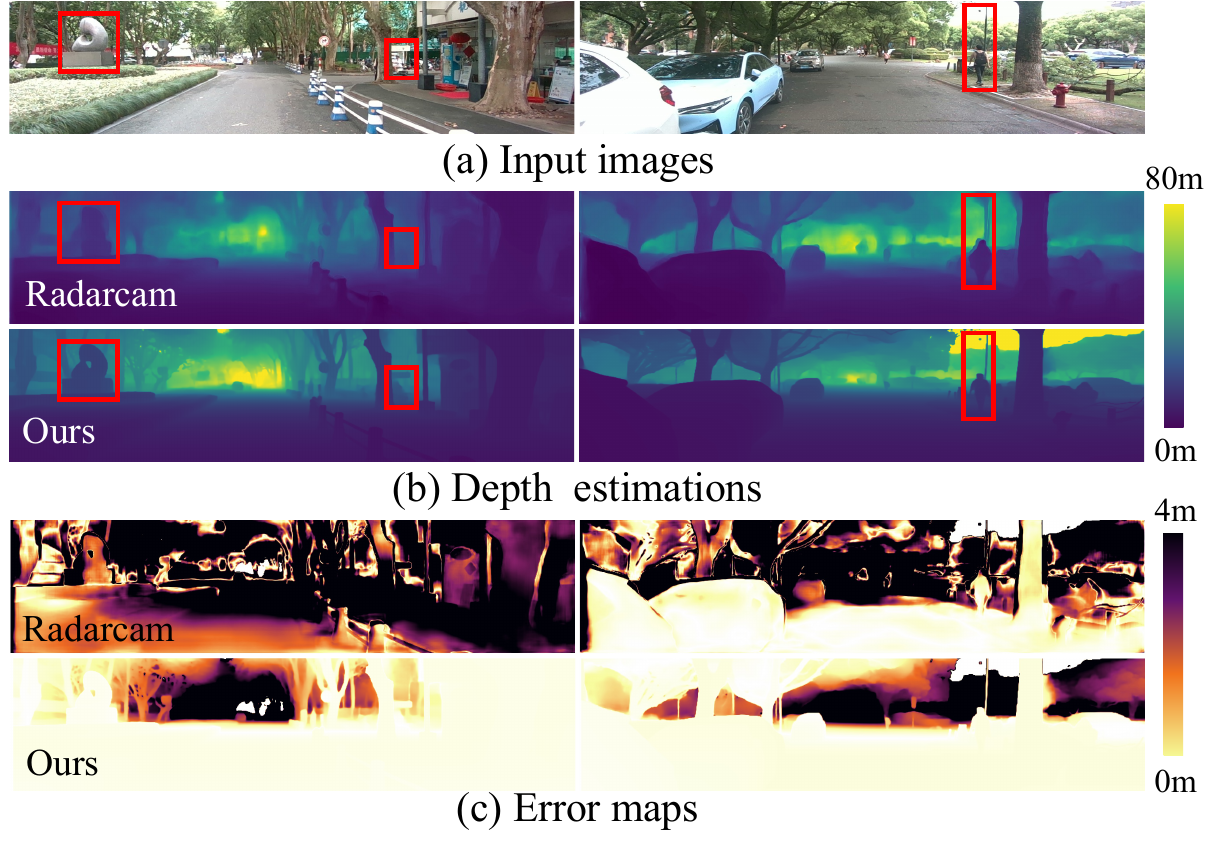} % 适当留白
    \caption{Experimental results for ZJU-4DRadarCam (from top to bottom): (a) Input image, (b) Radarcam's and our RaCalNet depth estimations, (c) Radarcam's and our errors. All values are in meters (m). Red boxes highlight superior performance regions.}
    \label{fig:zju}
\end{figure}

\begin{table}[!tbp]
\centering
\caption{EVALUATIONS ON ZJU-4DRADARCAM(CORRECTED) (mm)}
\label{tab:zju_corrected}
\small
\setlength{\tabcolsep}{5pt}
\resizebox{\columnwidth}{!}{%
\begin{tabular}{@{}llccccccc@{}}
\toprule
Dist & Method & MAE$\downarrow$ & RMSE$\downarrow$ & AbsRel$\downarrow$ & SqRel$\downarrow$ & $\delta_1\uparrow$ \\
\midrule
\multirow{2}{*}{50m} 
& Radarcam \cite{Radarcam2024} & 1468.645 & 3408.806 & 0.116 & 848.655 & 0.879 \\
& Ours & \textbf{1212.372} & \textbf{2137.652} & \textbf{0.095} & \textbf{535.929} & \textbf{0.895} \\
\midrule

\multirow{2}{*}{70m}
& Radarcam \cite{Radarcam2024} & 1572.917 & 3719.426 & 0.117 & 875.905 & 0.877 \\
& Ours & \textbf{1317.045} & \textbf{2358.958} & \textbf{0.097} & \textbf{572.155} & \textbf{0.893} \\
\midrule

\multirow{2}{*}{80m}
& Radarcam \cite{Radarcam2024} & 1603.922 & 3831.740 & 0.117 & 886.439 & 0.877 \\
& Ours & \textbf{1343.483} & \textbf{2479.008} & \textbf{0.098} & \textbf{583.625} & \textbf{0.892} \\
\bottomrule
\end{tabular}%
}
\end{table}

\begin{figure*}[!htbp] % 注意这里的星号*
    \centering
    \includegraphics[width=0.9\linewidth]{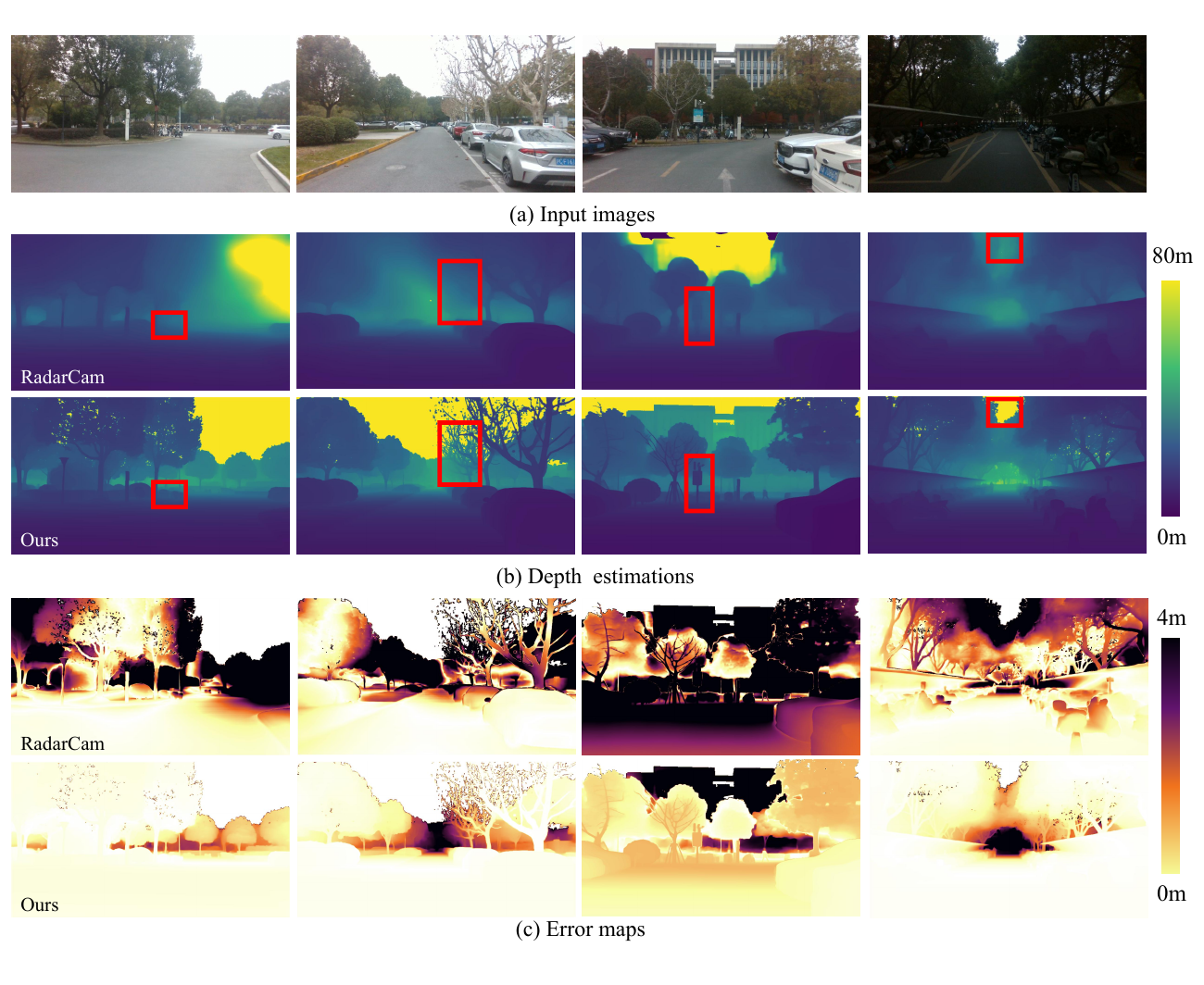} 
    \caption{Experimental results for proprietary real-world data (from top to bottom): (a) Input image, (b) Radarcam's and our RaCalNet depth estimations, (c) Radarcam's and our errors. All values are in meters (m). Red boxes highlight superior performance regions.}
    \label{fig:sjtu}
\end{figure*}

Fig.~\ref{fig:zju} presents visual comparisons organized in five rows. Red boxes highlight regions where RaCalNet exhibits noticeably better performance. The results show that our method accurately segments sky regions, delineates object boundaries with high precision, and effectively suppresses artifacts in complex scenes.

The superior performance of RaCalNet can be attributed to its two-stage architecture. The Radar Recalibration module filters out unreliable radar points and refines their projected positions, reducing noise in the sparse depth prior. Subsequently, the Metric Depth Optimization module aligns monocular predictions with the corrected sparse ground truth, enforcing global metric consistency. Together, these components allow RaCalNet to produce depth maps that are not only accurate in scale but also structurally coherent.

\subsection{Evaluation on Real-World Deployment Scenarios}

Following the same evaluation protocol as in Section~IV-C, we further validate the proposed framework using proprietary real-world deployment data. As summarized in Table~\ref{tab:sjtu}, our method consistently outperforms Radarcam across all tested distance ranges. In particular, compared to Radarcam, our method achieves substantial RMSE reductions of 46.24\%, 43.42\%, and 40.06\% at 50m, 70m, and 80m intervals, respectively. These results set a new benchmark for long-range radar-based dense depth completion.

Qualitative comparisons in Fig.~\ref{fig:sjtu} further illustrate the improvements in visual consistency, depth boundary clarity, and artifact suppression achieved by our method, particularly in challenging outdoor scenarios.

\begin{table}[!tbp]
\centering
\caption{EVALUATIONS ON REAL-WORLD DEPLOYMENT (mm)}
\label{tab:sjtu}
\small
\setlength{\tabcolsep}{5pt}
\resizebox{\columnwidth}{!}{%
\begin{tabular}{@{}llccccccc@{}}
\toprule
Dist & Method & MAE$\downarrow$ & RMSE$\downarrow$ & AbsRel$\downarrow$ & SqRel$\downarrow$ & $\delta_1\uparrow$ \\
\midrule
\multirow{2}{*}{50m} 
& Radarcam \cite{Radarcam2024} & 1469.553 & 3237.304 & 0.112 & 660.747 & 0.884 \\
& Ours & \textbf{1272.727} & \textbf{1982.943} & \textbf{0.091} & \textbf{376.293} & \textbf{0.905} \\
\midrule

\multirow{2}{*}{70m}
& Radarcam \cite{Radarcam2024} & 1674.472 & 3681.323 & 0.113 & 707.069 & 0.882 \\
& Ours & \textbf{1391.237} & \textbf{2340.489} & \textbf{0.094} & \textbf{425.510} & \textbf{0.899} \\
\midrule

\multirow{2}{*}{80m}
& Radarcam \cite{Radarcam2024} & 1721.122 & 3875.754 & 0.114 & 729.197 & 0.881 \\
& Ours & \textbf{1475.972} & \textbf{2523.246} & \textbf{0.096} & \textbf{446.519} & \textbf{0.897} \\
\bottomrule
\end{tabular}%
}
\end{table}

\subsection{Ablation Studies}

We conduct detailed ablation studies on the real-world deployment data to analyze the contribution of individual components in our framework. Specifically, we evaluate three configurations: (1) the complete RaCalNet pipeline; (2) a variant retaining radar screening but without Pixel-wise Displacement Refinement; and (3) a variant excluding the entire Radar Recalibration module (both screening and displacement refinement). Results are shown in Table~\ref{tab:ablation} and reveal the following insights:

\begin{table}[!tbp]
\centering
\caption{ABLATION OF OUR MODULE (mm)}
\label{tab:ablation}
\small
\setlength{\tabcolsep}{5pt}
\resizebox{\columnwidth}{!}{%
\begin{tabular}{@{}llccccccc@{}}
\toprule
Dist & Configuration & MAE$\downarrow$ & RMSE$\downarrow$ & AbsRel$\downarrow$ & SqRel$\downarrow$ & $\delta_1\uparrow$ \\
\midrule
\multirow{3}{*}{50m} 
& Complete RaCalNet & \textbf{1272.727} & \textbf{1982.943} & \textbf{0.091} & \textbf{376.293} & \textbf{0.905} \\
& w/o Displacement Refinement & 1326.122 & 2055.036 & 0.095 & 389.973 & 0.901 \\
& w/o Radar Refinement & 2505.368 & 3487.922 & 0.179 & 543.408 & 0.802 \\
\midrule

\multirow{3}{*}{70m}
& Complete RaCalNet & \textbf{1391.237} & \textbf{2340.489} & \textbf{0.094} & \textbf{425.510} & \textbf{0.899} \\
& w/o Displacement Refinement & 1476.192 & 2439.610 & 0.099 & 443.531 & 0.894 \\
& w/o Radar Refinement & 2774.884 & 4227.314 & 0.182 & 644.136 & 0.787 \\
\midrule

\multirow{3}{*}{80m}
& Complete RaCalNet & \textbf{1475.972} & \textbf{2523.246} & \textbf{0.096} & \textbf{446.519} & \textbf{0.897} \\
& w/o Displacement Refinement & 1572.627 & 2608.443 & 0.102 & 461.596 & 0.892 \\
& w/o Radar Refinement & 2833.835 & 4470.089 & 0.184 & 675.578 & 0.785 \\
\bottomrule
\end{tabular}%
}
\end{table}

1) Displacement Refinement Ablation: When radar screening is retained but pixel-level displacement refinement is disabled, we observe consistent performance degradation, with MAE increasing by 4.0-6.1\% across all depth ranges. This ablation highlights two key insights: (i) although unreliable radar points are removed through screening, the remaining points still suffer from slight misalignments caused by residual errors and temporal desynchronization between sensors, and (ii) the learned displacement refinement effectively compensates for these small yet impactful offsets, while preserving the geometric validity of radar projections.

2) Impact of Radar Refinement: The complete removal of the Radar Recalibration module (both screening and displacement components) leads to severe performance degradation, with MAE increasing by 96.8\% at 50m (2505mm vs. 1273mm) and 92.1\% at 80m (2834mm vs. 1476mm). This demonstrates that raw radar measurements contain substantial noise and errors that directly propagate to depth predictions. Our refinement module is crucial for: (i) identifying reliable radar points through cross-modal screening, and (ii) establishing accurate pixel-level correspondences, both being essential for maintaining metric accuracy in the fused output.

\subsection{Impact on Downstream Tasks: 3D Scene Reconstruction}

To demonstrate the practical benefits of high-quality depth estimation, we evaluate the influence of different depth prediction methods on a downstream 3D scene reconstruction task. Specifically, we employ FrozenRecon~\cite{Frozenrecon}, a 3D reconstruction framework that takes as input RGB images, depth maps, camera intrinsics, and optionally camera poses to produce dense 3D scene representations.

In our experiment, we supply the same RGB images and ground-truth camera intrinsics across all methods to ensure fair comparison. Camera poses are left to be optimized by FrozenRecon during reconstruction. The only varying input is the predicted depth map, which is either generated by the baseline method Radarcam~\cite{Radarcam2024} or produced by our complete RaCalNet pipeline.

\begin{figure*}[!htbp]
    \centering
    \includegraphics[width=0.9\textwidth]{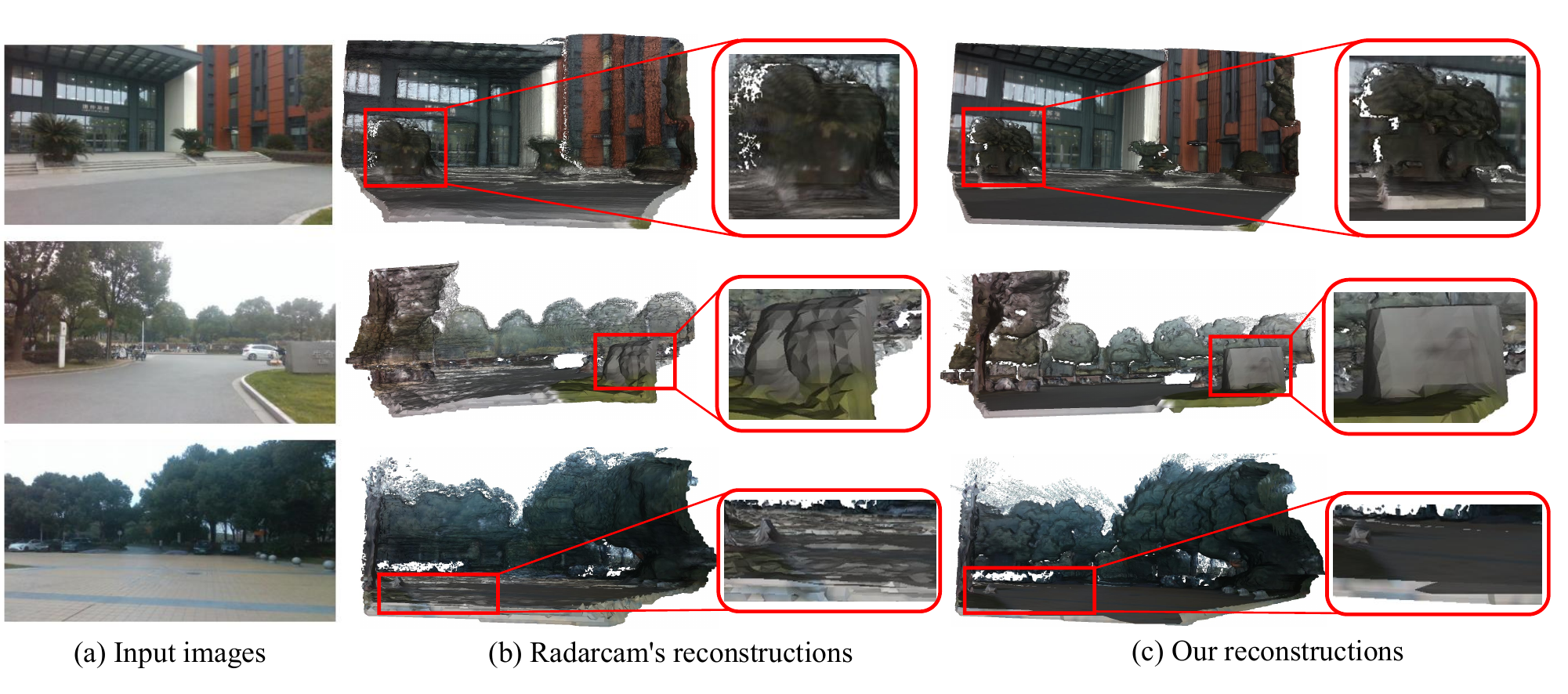}
    \caption{Qualitative comparison of 3D scene reconstruction results. (a) Input monocular RGB images. (b) Reconstructions using depth predicted by Radarcam, along with enlarged views of the highlighted regions. (c) Reconstructions based on depth estimated by our method, with corresponding enlarged views. Our approach yields more complete and geometrically accurate results.
}
    \label{fig:3drecon}
\end{figure*}

Fig.~\ref{fig:3drecon} shows qualitative comparisons of the reconstructed 3D scenes. Depth predictions from our method lead to noticeably more complete and geometrically accurate reconstructions, particularly in ground regions and structurally complex areas. These improvements underscore the broader utility of our framework, extending its value beyond standalone depth estimation to general 3D perception tasks.

\section{Conclusions}

We propose RaCalNet, a novel framework for dense depth estimation using radar and monocular images, which eliminates the need for dense LiDAR supervision. By leveraging sparse LiDAR to supervise the refinement of raw radar points, generating accurate, pixel-wise depth anchors. These anchors are then used to calibrate scale and optimize monocular depth predictions, addressing key challenges such as radar unreliability and the distortion introduced by dense LiDAR interpolation. Extensive experiments on public benchmarks and real-world scenarios demonstrate that RaCalNet achieves superior depth accuracy while using less than 1\% of the supervision density required by dense LiDAR-based methods. The resulting depth maps not only preserve fine-grained structural details but also enhance downstream tasks such as 3D reconstruction.

While RaCalNet is currently not directly applicable to 3D radar datasets such as nuScenes~\cite{nuscenes} due to the lack of elevation information and the resulting planar point clouds, this limitation can be addressed through elevation inference and multi-view consistency. In future work, we also plan to incorporate temporal modeling to further enhance robustness across sequential frames.

\newpage

\bibliographystyle{IEEEtran}
\bibliography{main}

\vfill

\end{document}